\documentclass[10pt,twocolumn,letterpaper]{article}

\usepackage[pagenumbers]{iccv} 
\usepackage{booktabs}
\usepackage{multirow}
\usepackage{amssymb}
\usepackage{pifont}
\usepackage{tcolorbox}
\usepackage{arydshln}
\usepackage{xcolor,colortbl,afterpage,capt-of,multicol}

\usepackage{amssymb}
\usepackage{pifont}

\usepackage{mathtools}

\definecolor{wino}{HTML}{cddefa}
\definecolor{dc4870}{HTML}{e0fccf}
\definecolor{genai}{HTML}{fae6ec}

\usepackage{siunitx}
\usepackage{numprint}

\definecolor{iccvblue}{rgb}{0.21,0.49,0.74}
\definecolor{myapricot}{RGB}{249,224,199}
\definecolor{mygray}{RGB}{238,238,238}
\usepackage[pagebackref,breaklinks,colorlinks,allcolors=iccvblue]
{hyperref}
\usepackage{tabularx}
\usepackage{multirow}
\usepackage{array}
\usepackage[table,xcdraw]{xcolor}
\usepackage{tcolorbox}
\def\confYear{2025}

\title{DT2IT-MRM: Debiased Preference Construction and Iterative Training for Multimodal Reward Modeling}

\author{Zhihong Zhang\thanks{University of Science and Technology of China}
\and
Jie Zhao
\and 
Xiaojian Huang
\and
Jin Xu
\and 
Zhuodong Luo
\and
Xin Liu
\and
Jiansheng Wei
\and 
Xuejin Chen
}

\begin{document}
\maketitle

\begin{abstract}
Multimodal reward models (MRMs) play a crucial role in aligning Multimodal Large Language Models (MLLMs) with human preferences. Training a good MRM requires high-quality multimodal preference data. However, existing preference datasets face three key challenges: lack of granularity in preference strength, textual style bias, and unreliable preference signals. Besides, existing open-source multimodal preference datasets suffer from substantial noise, yet there is a lack of effective and scalable curation methods to enhance their quality. To address these limitations, we propose \textbf{DT2IT-MRM}, which integrates a \textbf{D}ebiased preference construction pipeline, a novel reformulation of text-to-image (\textbf{T2I}) preference data, and an \textbf{I}terative \textbf{T}raining framework that curates existing multimodal preference datasets for \textbf{M}ultimodal \textbf{R}eward \textbf{M}odeling. Our experimental results show that DT2IT-MRM achieves new \textbf{state-of-the-art} overall performance on three major benchmarks: VL-RewardBench, Multimodal RewardBench, and MM-RLHF-RewardBench.
\end{abstract}

\section{Introduction}
\label{sec:intro}

Although Multimodal Large Language Models (MLLMs)~\cite{bai2025qwen2,Bai2025Qwen3VLTR,zhu2025internvl3,wang2025internvl3} have demonstrated promising performance in visual understanding and reasoning, they may occasionally generate suboptimal responses that are misaligned with human preferences.
Multimodal Reward Models (MRMs) have been studied to address these problems in both the training phase and the inference phase.
At the training phase, MRMs play a crucial role in post-training, including preference optimization and reinforcement learning (RL)~\cite{ouyang2022training,rafailov2023direct,sun2023aligning,xiong2024llava}. 
During preference optimization, MRM serves as a cost-effective and scalable alternative to manual annotation for constructing multimodal preference data. 
For RL, MRMs are particularly important in domains lacking verifiable answers~\cite{ji2025safe,zhou2025generative}. 
Compared to rule-based rewards~\cite{guo2025deepseek,meng2025mm}, MRMs are better suited for modeling complex human preferences required for general preference learning~\cite{liu2025skywork}.
At inference time, MRMs can be employed in various test-time scaling strategies (e.g., best-of-$N$) to identify the most optimal response among multiple candidates~\cite{wang2025visualprm,snell2024scaling}.

\begin{figure*}[t]
    \centering
    \includegraphics[width=1\linewidth]{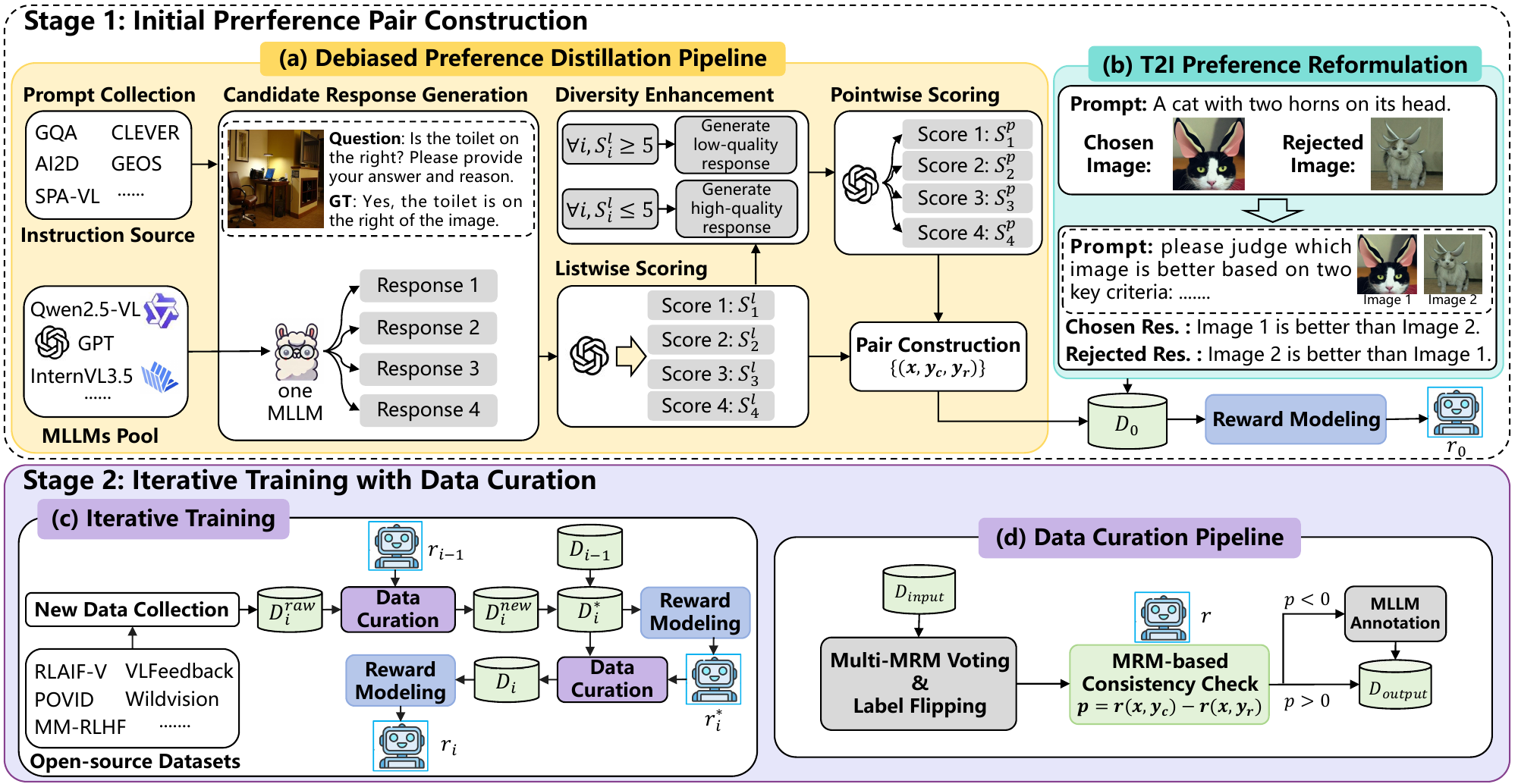}
    \caption{Overview of Our DT2IT-MRM. \textbf{In Stage 1-Preference Data Construction}, we construct single-image and multi-image preference data respectively for the initial MRM training. \textbf{(a)} Our debiased preference distillation pipeline for single-image preference pairs consists of five steps: (1) Prompt Collection, (2) Candidate Response Generation, (3) Listwise scoring, (4) Diversity Enhancement for diverse preferences, and (5) Pointwise Scoring to alleviate positional bias. \textbf{(b)} To construct multi-image preference data, we convert raw text-to-image preference data into multi-image, image-understanding preference data for easier learning by MRMs. \textbf{(c) In Stage 2 for MRM training}, we iteratively alternate between training MRMs and curating preference data. \textbf{(d)} Our data curation pipeline consists of two types of reward model consistency checks and MLLM annotations.
}
    \label{Fig:overview}
\end{figure*}

Training MRMs requires preference data to exhibit three essential characteristics: unbiasedness, diversity, and reliability. Existing multimodal preference datasets face several limitations. 
\textbf{1. Textual style bias}. 
MM-RLHF~\cite{zhang2025mm} generates multiple responses per prompt using a fixed set of models and ranks them manually to construct preference data. The chosen responses often come from higher-performing models, potentially introducing spurious correlations between response quality and text style. Consequently, MRMs may be biased to judge preferences based on text style rather than actual content quality. IXC-2.5-Reward~\cite{zang2025internlm} also exhibits text style bias in general and text-rich subsets because rejected responses are generated by a single model. 
\textbf{2. Lack of diversity in preference strength}. 
The preference data constructed by IXC-2.5-Reward~\cite{zang2025internlm} and InternVL2.5-MPO~\cite{wang2024enhancing} exhibit relatively uniform preference strength, with most preferences being strong and lacking variation across different levels of preference strength. Training MRMs with such data requires a large amount of data and tends to result in overfitting~\cite{wang2024secrets}. 
\textbf{3. Unreliable preference signals}.
Methods like VLFeedback~\cite{li2024vlfeedback}, LLaVA-Critic~\cite{xiong2024llava}, IXC-2.5-Reward, and Skywork-VL Reward~\cite{wang2025skywork} adopt pairwise or listwise evaluation, directly employing proprietary models to distill preference signals among multiple candidate responses without providing ground-truth reference answers. If the proprietary models cannot correctly answer the target question, their assessment of candidate responses becomes fundamentally unreliable. Moreover, closed-source models often suffer from positional bias~\cite{chen2024mllm,cheng2025evaluating} when evaluating multiple responses, which further compromises the reliability of the distilled preferences. 

To address the aforementioned data limitations, we propose a debiased preference distillation pipeline (\cref{Fig:overview}(a)).
To mitigate textual style bias, we use one MLLM to generate multiple candidate responses. To further address the issues of limited response and preference strength diversity, we design a diversity enhancement module. To ensure preference reliability, we combine listwise and pointwise scoring to address positional bias. Meanwhile, during the distillation process, we provide the ground-truth reference answer to the teacher model and guide it to produce structured reasoning.
Thus, our constructed preference data possess three key features: mitigated textual style bias, diverse preference strengths, and high reliability.
To further exploit more high-quality data and construct multi-image initial preference data, we transform existing large-scale human-annotated text-to-image preference data into a novel formulation for training discriminative MRMs. 

While automatically generating large-scale high-quality preference data is significantly challenging, some prior work, Skywork-VL Reward~\cite{wang2025skywork} attempts to curate noisy open-source data by directly using GPT-4o to re-distill preferences for all original open-source preference data, filtering out low-quality preferences. 
However, this method is prohibitively costly when applied to large-scale datasets, and its performance is also constrained by GPT-4o.
Thus, we design an effective and low-cost iterative training framework that progressively curates open-source data using MRMs and retrains the MRMs with the improved data, thus jointly improving both the MRM and data quality, without being constrained by proprietary models.
Our data curation pipeline (\cref{Fig:overview}(d)) utilizes two kinds of reward model consistency checks and a small amount of MLLM annotation. 
Our method significantly reduces the need for MLLM annotations.

In summary, our \textbf{DT2IT-MRM} integrates the \textbf{D}ebiased preference distillation pipeline, the text-to-image (\textbf{T2I}) preference reformulation, and the \textbf{I}terative \textbf{T}raining framework.
Compared to the previous state-of-the-art (SOTA) method BaseReward (75.2\%)~\cite{zhang2025basereward}, our DT2IT-MRM achieves a new SOTA overall accuracy of 80.5\% on three major benchmarks: VL-RewardBench~\cite{li2025vl}, Multimodal RewardBench~\cite{yasunaga2025multimodal}, and MM-RLHF-RewardBench~\cite{zhang2025mm}, while using only approximately 35\% of the data employed by BaseReward.
On VL-RewardBench, our DT2IT-MRM achieves the best performance with an overall accuracy of 83.5\%, surpassing BaseReward’s 82.2\%. Notably, on the large-scale Multimodal RewardBench, our method yields an improvement of approximately 9\% over the previous SOTA. 
These results demonstrate the data efficiency and strong generalization capability of our method.

Our main contributions are summarized as follows:\\
• We propose an \textbf{automatic construction pipeline} that systematically mitigates textual style bias and positional bias to generate \textbf{debiased} preference data for MRM training.\\
• We introduce a \textbf{novel reformulation of text-to-image preference data}, which is more effective for training discriminative MRMs.\\
• We develop an \textbf{iterative training framework} that jointly performs MRM training and automatic preference data curation, progressively improving both model performance and data quality.\\
• We construct and curate a \textbf{large-scale, high-quality multimodal preference dataset} consisting of \textbf{929K} preference pairs, covering both \textbf{single-image} and \textbf{multi-image} inputs.
\section{Related Work}
\label{sec:related_work}

\subsection{Multimodal Preference Data Construction} 
Existing multimodal preference construction approaches include human annotation, AI-generated feedback, and rule-based techniques. 
Approaches that rely on human annotation~\cite{zhang2025mm,wang2025skywork} are time-consuming and costly, with limited scalability, and are susceptible to inconsistencies across annotators.
Methods that use AI-generated feedback, such as VLFeedback~\cite{li2024vlfeedback}, LLaVA-Critic~\cite{xiong2024llava}, and IXC-2.5-Reward~\cite{zang2025internlm}, directly use closed-source models to distill preferences. 
However, the absence of ground truth reference answers during distillation, along with the positional bias or text style bias in these methods, leads to low-quality preferences.
RLAIF-V~\cite{yu2025rlaif} proposes a divide-and-conquer framework to build feedbacks using open-source MLLMs.
However, the limited capabilities of these open-source models~\cite{liu2024llavanext,omnilmm} affect data quality. 
For rule-based techniques, several works~\cite{zang2025internlm,wang2024enhancing} construct preferences by verifying the correctness of the final answer. However, they overlook the correctness of the intermediate reasoning in the responses, which may lead to unreliable preferences.
Other methods~\cite{zhou2024aligning,pi2025mr} generate preferences by injecting errors, but this data suffers from poor generalization, as MRMs learn the inherent patterns between edited and original responses~\cite{li2025devil}.
To construct diverse, debiased and reliable preferences data, we propose a new data construction pipeline.

\subsection{Multimodal Reward Models}
Existing MRMs can be classified into three categories: generative, semi-scalar, and discriminative.
\textbf{Generative MRMs} directly employ a MLLM as the MRM to generate textual critiques. Existing works typically train such MRMs using supervised fine-tuning (SFT)~\cite{xiong2024llava,wang2025unified2}, reinforcement learning (RL)~\cite{wang2025llava}, or combining SFT and RL~\cite{zhang2025r1,wang2025unified,pi2025mr}.
However, these MRMs heavily rely on the MLLM’s inherent capabilities of understanding and reasoning.
\textbf{Semi-scalar MRMs}~\cite{zhang2025mm} generate both textual critiques and a scalar reward score.
Its performance depends heavily on critique quality, and despite being trained on human-annotated data, it still fails to achieve satisfactory results.
\textbf{Discriminative MRMs}~\cite{zang2025internlm,wang2025skywork,zhang2025basereward,jin2025omni} connect a reward head to the MLLM's base model, directly outputting a scalar reward score for the response. Despite limited interpretability, they excel in simplicity and are computationally efficient.
Our DT2IT-MRM belongs to the discriminative MRMs.
\section{Method}
\label{sec:method}

To improve the quality of preference data for training MRMs, we propose \textbf{DT2IT-MRM}, which comprises a \textbf{D}ebiased preference construction pipeline, a text-to-image (\textbf{T2I}) reformulation method, and an \textbf{I}terative \textbf{T}raining framework.
First, to generate high-quality initial preference data, we design two data construction methods: (1) a debiased preference distillation pipeline (\cref{sec:pipline}), and (2) a new text-to-image preference reformulation (\cref{sec:reformulation}).
Second, our iterative training framework (\cref{sec:progressive_learning}) progressively curates open-source data using the MRMs and retrains the MRMs with the improved data, thus jointly improving both the MRM and training data quality.

\subsection{Debiased Preference Distillation}
\label{sec:pipline}
Training discriminative multimodal reward models requires multimodal preference data in the form of triplets: (image-text prompt $x$, chosen response $y_c$, rejected response $y_r$). To automatically generate large-scale triplets without introducing textual style and positional bias, we design a novel five-step preference data distillation pipeline (\Cref{Fig:overview}(a)).

\textbf{Step 1: Prompt collection.}
We first collect multimodal prompts from supervised fine-tuning datasets, covering visual understanding, visual reasoning, and multimodal safety, with each prompt $x$ paired with a corresponding reference answer $\hat{y}$. Additional details can be found in the appendix.

\textbf{Step 2: Candidate response generation.} To generate multiple candidate responses, we first construct a pool of MLLMs of various capabilities: GPT-5-chat~\cite{gpt5}, GPT-4o-mini~\cite{gpt4omini}, Qwen2.5-VL (7B, 32B)~\cite{bai2025qwen2}, and InternVL3.5 (8B, 38B)~\cite{wang2025internvl3}. 
For each prompt $x$, 
we randomly select a MLLM from the pool to generate $K$ responses ($\{ y_i \}_{i=1}^K$) using a high temperature and nucleus sampling parameters to encourage response diversity.
Using the same model ensures that the chosen and rejected responses are sampled from the same distribution, thereby sharing similar textual styles and linguistic patterns. 
This operation alleviates the potential spurious correlation between text styles and response quality, thereby effectively mitigating textual style bias.

\textbf{Step 3: Listwise scoring.} For each prompt $x$, we employ GPT-5.2 to score the generated $K$ candidate responses $\{ y_i \}_{i=1}^K$ concurrently, given the ground-truth reference answer $\hat{y}$ obtained at Step 1.
To enhance the reliability of the scoring process, we first establish a comprehensive set of evaluation criteria, encompassing $M$ aspects ($M=6$ in this work) from existing works: accuracy, helpfulness, completeness, language quality, creativity, and ethical consideration. 
We then design a structured reasoning process for GPT-5.2 to generate reliable scores considering the evaluation criteria. 
First, we let GPT-5.2 generate its own reference answer $\tilde{y}$, then predicts the importance weights $\{w_i^{m}\}_{m=1}^M$ of each $y_i$ for each evaluation criterion.
Second, we employ GPT-5.2 to perform textual analysis and generate scores $\{s_i^m\}\in \{0,1,\ldots,10\}$ on each evaluation criterion.
Finally, the overall score $S_i^l$ for each response $y_i$ is computed as a weighted average of the scores across all evaluation criteria:
$S_i^l=\sum_{m=1}^{M} w_i^m s_i^m$. 
The detailed evaluation prompts can be found in the appendix.

\textbf{Step 4: Response diversity enhancement.}
When generating multiple candidate responses to the same prompt in Step 2, it may result in multiple responses that are either all high-quality or all low-quality, which hinders the preference construction and reduces the utility of the data. 
After listwise scoring, if all generated responses $\{y_i\}^{K}_{i=1}$ receive low scores, we randomly select a model from multiple proprietary models to generate a better response with access to the reference answer. 
Note that we do not use one identical proprietary model for all such cases to ensure diverse textual styles of the generated higher-quality responses.
Conversely, if all generated responses have high scores, we inject noise into the input image and generate lower-quality responses from the noised input with the previously selected MLLM, thus avoiding text style bias for all the $K$ responses.

\textbf{Step 5: Pointwise scoring.} 
The listwise scoring mechanism in step 3 will inevitably introduce positional bias~\cite{chen2024mllm,cheng2025evaluating}. 
To alleviate the positional bias, we combine listwise scoring with pointwise scoring.
In pointwise scoring, each candidate response to a given prompt is scored ($S_i^p$) using GPT-5.2 independently using the same weighted scoring process as in Step 3, without comparing to other responses. 
After applying both listwise and pointwise scoring, each candidate response $y_i$ to a given prompt is assigned two scores: $S_i^l$ and $S_i^p$. 

We then construct preference pairs based on the two scores, respectively.
Considering the listwise scores, for any pair of response $y_i$ and $y_j$ whose $S^{l}_i>S^{l}_j$, $y_i$ is selected as the chosen response and $y_j$ as the rejected response.
Same operation is performed for the pointwise scores. 
Ultimately, we retain only the preference pairs $(x,y_i,y_j)$ that satisfy both $S_i^l>S_j^l$ and $S_i^p>S_j^p$, and add them to our initial preference dataset $D_0$.

\subsection{Text-to-Image Preference Reformulation}
\label{sec:reformulation}
While our debiased preference distillation pipeline generates preference pairs for single-image understanding, we further develop a method for generating multi-image preference pairs. 
We exploit existing large-scale human-annotated text-to-image preference datasets~\cite{han2024evalmuse,wu2023human} that contain preference data (text prompt $x$, chosen image $I_c$, rejected image $I_r$) for training text-to-image models.
However, these triplets can not be directly applied to train MRMs, since existing MRMs are based on MLLMs which only generate textual outputs.
To make the data applicable to MRMs, Omni-Reward~\cite{jin2025omni} swaps the order of text and images to finetune MRMs. Specifically, the chosen or rejected image is directly used as model input, while the text prompt $x$ serves as the model output. 
To train discriminative MRMs, the Bradley-Terry style loss function~\cite{ouyang2022training} is typically used given a multimodal preference pair (image-text prompt $x'$, chosen response $y_c$, rejected response $y_r$):
\begin{equation}
L=-E[\log(\sigma(r(x',y_c)-r(x',y_r)))],
\label{eq:origin_loss}
\end{equation}
where $\sigma$ is the logistic function, and $r$ denotes the MRMs.
Omni-Reward modifies the loss function to fit the text-to-image preference pairs as follows: 
\begin{equation}
L=-E[\log(\sigma(r(I_c,x)-r(I_r,x)))].
\label{eq:omnireward}
\end{equation} 
To better align with the image understanding capabilities of MLLMs and facilitate MRM training, we propose a new reformulation method that transforms the text-to-image task into a pairwise evaluation task for two images. 
The model input now consists of the chosen image $I_c$, rejected image $I_r$, text prompt $x$, and evaluation prompt $E$. The model output is a textual evaluation result, such as ``Image 1 is better than Image 2", which is used as the chosen ($y_c$) or rejected response ($y_r$). The reformulated preference data is represented as: $( (I_c, I_r, x, E), y_c, y_r)$. The order of the chosen and rejected images is randomly shuffled in the model input. The loss function is then revised as follows:
\begin{equation}
\begin{array}{c}
L = -E\left[\log\left(\sigma\left(r(X,y_c) - r(X,y_r)\right)\right)\right], \\
X = (I_c, I_r, x, E)\ \ or\ \ (I_r,I_c,x,E).
\end{array}
\label{eq:ourform}
\end{equation}

\subsection{Iterative Training Framework}
\label{sec:progressive_learning}
Although we train a MRM $r_0$ using the initial preference data $D_0$ constructed in \cref{sec:pipline,sec:reformulation}, $D_0$ remains relatively small-scale and contains noise. To leverage larger-scale open-source data, which contain even more noise, we aim to use MRMs to filter out the noise. However, this requires stronger MRMs. Thus, we propose an iterative training framework that progressively curates data using MRMs and retrains the MRMs on the curated data, jointly improving both data quality and model performance.

To curate noisy preference datasets, our data curation pipeline consists of three steps. The \textbf{first step} primarily focuses on correcting noise within open-source datasets. Following previous work~\cite{wang2024secrets}, for each dataset, we use a multi-MRM voting method to estimate the preference strength of each preference pair. We then sort all the data in ascending order of preference strength. For data with negative preference strength, we flip the preference labels of the bottom 50\% of these data.
In the \textbf{second step}, we use our trained MRM to conduct preference consistency filtering. For each preference pair processed after the first step, if the preference label is consistent with our trained MRM (i.e., the reward score for the chosen response is higher than that for the rejected response), the data is retained. Otherwise, it proceeds to the third step.
In the \textbf{third step}, for data inconsistent with our trained MRM, we ensemble multiple MLLMs to re-annotate these preferences through pairwise evaluation. To mitigate positional bias, each model performs pairwise evaluations twice by swapping the response order. 
The final preference label is determined by the majority of the voting results. In case of a tie, the preference pair is discarded.

Our iterative training framework consists of five main steps. The initial MRM $r_0$ is trained using our initial preference data $D_0$.
For the $i$-th iteration, we first collect open-source multimodal preference datasets $D_i^{raw}$.
Second, we curate $D_i^{raw}$ using the last iteration's MRM $r_{i-1}$ with our data curation pipeline to obtain $D_i^{new}$.
Third, we combine the curated open-source data ($D_i^{new}$) with the training data from the last iteration ($D_{i-1}$) to obtain $D_i^*$, and train a better-performing MRM ($r_i^*$) on $D_i^*$.
Fourth, to further eliminate noise in the existing training data, we reapply our data curation pipeline to $D_i^*$ using $r_i^*$ to obtain $D_i$, while skipping the first step of the curation pipeline.
Finally, we train the MRM ($r_i$) on $D_i$. These five steps are performed iteratively.

\section{Experiments}

\begin{table*}[!ht]
\caption{Overall performance comparison of our DT2IT-MRM with top-performing multimodal reward models on three major benchmarks.}
\label{overall_preformance}
\centering
\setlength{\tabcolsep}{0.75mm}
\renewcommand{\arraystretch}{0.95}
\begin{small}
\begin{tabular}{lccccc}
\toprule
Models & VL-RewardBench & Multimodal RewardBench & MM-RLHF-RewardBench & \textbf{Mean Acc} & \textbf{Overall Acc} \\
\midrule
\#Samples & 1247 &  4711 & 170 & 6128 & 6128  \\
\midrule
GPT-5.2  &  70.2 & \underline{75.3} & 68.2 & 71.2 & 74.1  \\
Skywork-VL Reward & 73.1 & 74.4 & 65.9 & 71.1 & 73.9 \\
BaseReward  &  \underline{82.2} & 72.8    & \textbf{91.8}   &  \underline{82.2}   &   \underline{75.2}    \\
DT2IT-MRM (Ours)  &  \textbf{83.5}  & \textbf{79.3}   & \underline{89.4}  &  \textbf{84.1}  & \textbf{80.5} \\
\bottomrule
\end{tabular}
\end{small}
\end{table*}

\begin{table*}[!ht]
\caption{Performance comparison of our DT2IT-MRM with three categories of multimodal reward models on \textbf{VL-RewardBench}. * denotes our evaluation results with their released model.}
\label{vlrewardbench}
\centering
\renewcommand{\arraystretch}{0.95}
\begin{small}
\begin{tabular}{lcccccc}
\toprule
\textbf{Models}      & \textbf{\#Param} & \textbf{General} & \textbf{Hallucination} & \textbf{Reasoning} & \textbf{Macro Acc} & \textbf{Overall Acc} \\
\midrule
\rowcolor{orange!20}
\multicolumn{7}{l}{\textit{\textbf{Generative Multimodal Reward Models}}} \\
\midrule
\multicolumn{7}{c}{\textit{Proprietary Models (w/o critic training)}} \\
Claude-3.5-Sonnet (2024-06-22)  & -  & 43.4 & 55.0 & 62.3 & 53.6 & 55.3 \\
GPT-4o (2024-08-06) & - & 49.1 & 67.6 & 70.5 & 62.4  & 65.8\\
Claude-3.7-Sonnet    & -      & 68.1    & 70.7          & 60.8      & 66.5     &   66.3       \\
Gemini-1.5-Pro (2024-09-24)   & -      & 50.8    &  72.5       & 64.2      &    62.5       & 67.2    \\
GPT-5.2$^*$ & - & 52.5 & 71.8 & \textbf{76.7} & 67.0 & 70.2  \\
\midrule
\multicolumn{7}{c}{\textit{Open-Source Models (w/o critic training)}} \\
Qwen2.5-VL-7B-Instruct     & 7B      & 43.4    & 42.0      & 63.0     &      49.5     &  48.0 \\
InternVL3-8B     & 8B      & 60.6    & 44.0     & 62.3    &   55.6     &  57.0 \\
Qwen2.5-VL-72B-Instruct  & 72B     & 47.8    & 46.8          & 63.5     &   52.7      &   51.6   \\
Llama-3.2-90B        & 90B     & 42.6    & 57.3          & 61.7      &    53.9    &  56.2      \\
InternVL3-78B     & 78B     & 67.8    & 52.5          & 64.5     &    61.6    &  63.3    \\
\midrule
\multicolumn{7}{c}{\textit{Fast-Thinking Generative MRMs (with critic training)}} \\
LLaVA-Critic & 8B & 54.6 & 38.3 & 59.1 & 44.0 & 41.2 \\
UnifiedReward$^*$ & 7B & 45.9 & 66.6 & 64.0 & 58.8  & 63.0 \\
\midrule
\multicolumn{7}{c}{\textit{Slow-Thinking Generative MRMs (with critic training)}} \\
R1-Reward$^*$ & 7B & 43.1 & 57.5 & 39.1 & 46.6 & 50.8 \\
UnifiedReward-Think$^*$ &  7B & 51.4 & 82.9 & 61.8  & 65.4  &  73.0 \\
MR. Judge-7B-SFT-RL & 7B & \underline{68.7} & 83.2 & 61.4 & 71.1 & 75.5 \\
\midrule
\rowcolor{orange!10}
\multicolumn{7}{l}{\textit{\textbf{Semi-Scalar Multimodal Reward Models}}} \\
\midrule
MM-RLHF-Reward       & 7B      & 45.0    & 50.5          & 57.6      &      51.0    &   50.2   \\
\midrule
\rowcolor{orange!20}
\multicolumn{7}{l}{\textit{\textbf{Discriminative Multimodal Reward Models}}} \\
\midrule
IXC-2.5-Reward       & 7B      & \textbf{84.7}   & 62.5   & 62.9 & 70.0       & 65.8     \\
Omni-RewardModel-BT$^*$ & 8B & 51.4 & 81.4 & 61.5 & 64.8 & 72.0 \\
Skywork-VL Reward & 7B &  66.0 & 80.0 & 61.0  & 69.0 & 73.1 \\
\rowcolor{red!10}
BaseReward (Qwen2.5-VL) & 7B & 68.6 & 92.2 & 81.8 & 80.9 & 82.2 \\
\midrule
\rowcolor{orange!20}
\multicolumn{7}{l}{\textit{\textbf{Ours}}} \\
\midrule
DT2IT-MRM (Qwen2.5-VL)  &  7B      & 63.5    & \textbf{91.3}         &  \underline{72.6}      & \underline{75.8}      & \underline{82.5} \\
DT2IT-MRM (Qwen3-VL)  &  8B  & 63.0   & \textbf{91.3}  &  \textbf{76.7}  & \textbf{77.0}        & \textbf{83.5} \\
\bottomrule
\end{tabular}
\end{small}
\end{table*}

\begin{table*}[!ht]
\caption{Performance comparison of our DT2IT-MRM with three categories of multimodal reward models on \textbf{Multimodal RewardBench}.
}
\centering
\label{multimodal_rewardbench}
\setlength{\tabcolsep}{1mm}
\renewcommand{\arraystretch}{0.95}
\begin{small}
\begin{tabular}{lccccccccc}
\toprule
\multirow{2}{*}{\textbf{Models}} & \multirow{2}{*}{\textbf{\#Param}} & \multirow{2}{*}{\textbf{Overall}} & \multicolumn{2}{c}{\textbf{General}} & \multirow{2}{*}{\textbf{Knowledge}} & \multicolumn{2}{c}{\textbf{Reasoning}} & \multirow{2}{*}{\textbf{Safety}} & \multirow{2}{*}{\textbf{VQA}} \\
\cline{4-5} \cline{7-8}
& & & \textbf{Correctness} & \textbf{Preference} & & \textbf{Math} & \textbf{Coding} & & \\
\midrule
\rowcolor{orange!20}
\multicolumn{10}{l}{\textit{\textbf{Generative Multimodal Reward Models}}} \\
\midrule
\multicolumn{10}{c}{\textit{Proprietary Models (w/o critic training)}} \\
GPT-4o & - & 70.8 & 62.6 & \underline{69.0} & 72.0 & 67.6 & 62.1 & 74.8 & 87.2 \\
Claude-3.5-Sonnet & - & 71.5 & 62.6 & 67.8 & 73.9 & 68.6 & \underline{65.1} & 76.8 & 85.6 \\
Gemini-1.5-Pro & - & 71.9 & 63.5 & 67.7 & 66.3 & 68.9 & 55.5 & 94.5 & 87.2 \\
Claude-3.7-Sonnet & - & 71.9 & 58.4 & 60.7 & \textbf{78.1} & 76.3 & \textbf{71.3} & 72.0 & 86.8 \\
GPT-5.2$^*$ & - & 75.3 & 60.7 & 57.5 & \underline{77.5} & 78.2 & 62.5 & \textbf{99.8} & 85.9 \\
\midrule
\multicolumn{10}{c}{\textit{Open-Source Models (w/o critic training)}} \\
Llama-3.2-Vision-Instruct & 11B & 51.2 & 57.8 & 65.8 & 55.5 & 50.6 & 51.7 & 20.9 & 55.8 \\
Llama-3.2-Vision-Instruct & 90B & 61.2 & 60.0 & 68.4 & 61.2 & 56.3 & 53.1 & 52.0 & 77.1 \\
InternVL3-8B & 8B & 63.6 & 59.6 & 61.6 & 60.5 & 65.1 & 56.6 & 59.3 & 82.3 \\
Qwen2-VL-72B-Instruct & 72B & 70.9 & 56.4 & 62.3 & 70.2 & 73.3 & 58.9 & 90.1 & 85.3 \\
\midrule
\multicolumn{10}{c}{\textit{Fast-Thinking Generative MRMs (with critic training)}} \\
LLaVA-Critic$^*$ & 8B & 48.9 & 60.7 & 63.5 & 47.1  & 53.5 & 42.6  & 58.1 & 33.2 \\
UnifiedReward$^*$ & 7B & 65.7 & 65.7 & 60.7 & 56.2 & 59.9 & 50.0 & 88.6 & 73.7 \\
\midrule
\multicolumn{10}{c}{\textit{Slow-Thinking Generative MRMs (with critic training)}} \\
R1-Reward$^*$ & 7B & 53.0 & 47.0 & 45.0 & 43.0 & 57.2 & 27.5 & 80.9  & 64.3 \\
UnifiedReward-Think$^*$ &  7B & 62.1 & 65.2 & 63.5 & 58.1 & 61.5 & 51.5 & 45.9 & 74.0   \\
\midrule
\rowcolor{orange!20}
\multicolumn{10}{l}{\textit{\textbf{Semi-Scalar Multimodal Reward Models}}} \\
\midrule
MM-RLHF-Reward & 7B & 67.1 & 61.7 & 67.5 & 54.3 & 58.4 & 57.9 & 92.9 & 76.8 \\
\midrule
\rowcolor{orange!20}
\multicolumn{10}{l}{\textit{\textbf{Discriminative Multimodal Reward Models}}} \\
\midrule
IXC-2.5-Reward & 7B & 66.6 & 60.7 & 64.2 & 56.8 & 63.0 & 50.5 & 89.9 & 81.1 \\
Omni-RewardModel-BT & 8B & 70.5 &71.3 & 58.4  & 66.7 & 71.0 & 48.5 & 79.3 & 85.1 \\
BaseReward (Qwen2.5-VL) & 7B & 72.8 & 65.7 & 65.0& 70.6 &82.7 & 50.3 & 81.5 & 85.0  \\
Skywork-VL Reward$^*$ & 7B & 74.4 & 67.9 & 68.5 & 67.0 & 70.6  & 62.0 & 91.3 & 85.3 \\
\midrule
\rowcolor{orange!20}
\multicolumn{10}{l}{\textit{\textbf{Ours}}} \\
\midrule
DT2IT-MRM (Qwen2.5-VL) & 7B & \underline{77.7} & \underline{77.7} & 68.5 & 71.0 & \underline{84.4} & 52.9 & \underline{96.9} & \underline{87.3}  \\
DT2IT-MRM (Qwen3-VL) & 8B & \textbf{79.3} & \textbf{79.5} & \textbf{71.6} & 75.7 & \textbf{87.0} & 52.6 & 93.7 & \textbf{89.1}   \\
\bottomrule
\end{tabular}
\end{small}
\end{table*}

\begin{table*}[!ht]
\caption{Performance comparison of our DT2IT-MRM with three categories of multimodal reward models on \textbf{MM-RLHF-RewardBench}.}
\label{mmrlhf_rewardbench}
\centering
\renewcommand{\arraystretch}{0.95}
\begin{small}
\begin{tabular}{lcccccccc}
\toprule
\textbf{Models}      & \textbf{\#Param} & \textbf{Mcq} & \textbf{Long} & \textbf{Short} & \textbf{Safety} & \textbf{Video} & \textbf{Acc} & \textbf{Acc+} \\
\midrule
\rowcolor{orange!20}
\multicolumn{9}{l}{\textit{\textbf{Generative Multimodal Reward Models}}} \\
\midrule
\multicolumn{9}{c}{\textit{Proprietary Models (w/o critic training)}} \\
Gemini-2.0-Flash-Exp & - &  33.33 & 45.94  & 67.64 & 43.75 & 32.00 & 44.71 & 13.04 \\
GPT-4o (2024-08-06) & - & 64.28 & 78.37 & 44.11 & 56.25 & 40.00 & 58.23  & 26.01 \\
Claude-3.5-Sonnet (2024-06-22)  & -  & 64.28 & 67.56 & 55.88 & 65.62 &  60.00 & 62.94 & 26.11 \\
Claude-3.7-Sonnet    & -     & 66.67 & 91.89 & 91.18 & \textbf{87.50} & 76.00 & 82.35 & 65.22  \\
GPT-5.2$^*$ & - & 57.14 & 91.89 & 55.88 & 59.38 & 80.00 & 68.24 & 41.30 \\
\midrule
\multicolumn{9}{c}{\textit{Open-Source Models (w/o critic training)}} \\
InternVL3-8B     & 8B      &  35.71 & 56.76 & 23.53 & 37.50 & 32.00 & 37.65 & 6.52 \\
NVLM-D  & 72B     & 42.85 & 32.43 & 8.82 & 50.00 & 40.00 & 34.70  & 6.52  \\
Llama-3.2-90B        & 90B     & 19.04 & 35.13 & 38.23 & 50.00 & 40.00 & 35.29 & 10.86     \\
Qwen2-VL-72B-Instruct     & 72B     & 45.23 & 62.16 & 47.05 & 46.88 & 36.00 & 48.23 & 13.04  \\
\midrule
\multicolumn{9}{c}{\textit{Fast-Thinking Generative MRMs (with critic training)}} \\
LLaVA-Critic$^*$ & 8B & 66.67 & 81.08 & 64.71 & 71.88 & \textbf{92.00} & 74.12 & 47.83 \\
UnifiedReward$^*$ & 7B & 69.05 & 94.60 & 67.65 & \underline{84.38}  & 84.00 & 79.41 & 52.17 \\
\midrule
\multicolumn{9}{c}{\textit{Slow-Thinking Generative MRMs (with critic training)}} \\
R1-Reward$^*$ & 7B & 52.38  & 48.65 & 47.06 & 43.75 & 44.00 & 47.65 & 21.74 \\
UnifiedReward-Think$^*$ &  7B & 73.81 & 86.49 & 61.77  & 59.38  &  72.00 & 71.18 & 39.13 \\
\midrule
\rowcolor{orange!20}
\multicolumn{9}{l}{\textit{\textbf{Semi-Scalar Multimodal Reward Models}}} \\
\midrule
MM-RLHF-Reward       & 7B      & 83.00 & 97.00 & 74.00 & 69.00 & 88.00 & 82.00 &63.00 \\
\midrule
\rowcolor{orange!20}
\multicolumn{9}{l}{\textit{\textbf{Discriminative Multimodal Reward Models}}} \\
\midrule
Omni-RewardModel-BT$^*$ & 8B & 40.48 & 78.38 & 70.59 & 46.88 & 76.00 & 61.18 & 39.13 \\
Skywork-VL Reward$^*$ & 7B &  42.86 & 81.08 & 73.53  & 65.63 & 72.00 & 65.88 & 41.30 \\
IXC-2.5-Reward       & 7B      & 52.38 & 91.89 & 67.65 & 62.50 & \underline{88.00} & 71.18 & 50.00 \\
BaseReward (Qwen2.5-VL) & 7B & \textbf{95.74} &  \textbf{97.38} & 94.13 & 81.25 & \underline{88.00} & \textbf{91.76} & \textbf{80.43} \\
\midrule
\rowcolor{orange!20}
\multicolumn{9}{l}{\textit{\textbf{Ours}}} \\
\midrule
DT2IT-MRM (Qwen2.5-VL)  &  7B      & \underline{95.24}    & \underline{97.30} &  \textbf{100.00} & 71.88  & \underline{88.00} & \underline{91.18} & \underline{78.26}  \\
DT2IT-MRM (Qwen3-VL)  &  8B  & 90.48   & 94.60  &  \underline{97.06}  & 78.13 & 84.00 &  89.41 & 71.74   \\
\bottomrule
\end{tabular}
\end{small}
\end{table*}

\subsection{Evaluation Benchmarks and Metrics} 
There are three primary multimodal reward benchmarks, including VL-RewardBench~\cite{li2025vl}, Multimodal RewardBench~\cite{yasunaga2025multimodal}, and MM-RLHF-RewardBench~\cite{zhang2025mm}. 
VL-RewardBench~\cite{li2025vl} consists of 1,247 samples and uses two metrics: \textit{Overall Accuracy}, which measures the proportion of the model’s preference judgments consistent with human preferences, and \textit{Macro Average Accuracy}, which calculates the average accuracy across various tasks. 
Multimodal RewardBench~\cite{yasunaga2025multimodal} releases 4,711 samples and offers holistic evaluation across six aspects: general correctness, preference, knowledge, reasoning, safety, and visual question-answering.
MM-RLHF-RewardBench~\cite{zhang2025mm} contains 170 samples and uses two metrics: \textit{Traditional Accuracy (Acc)}, which measures the fraction of cases where the chosen response is correctly identified, and \textit{Acc+}, which measures the proportion of cases where all response pairs for a given multimodal prompt are correctly ranked. 
\textit{Acc+} focuses on evaluating the model's ability to judge weak preferences.

\subsection{Baseline Multimodal Reward Models}
We perform a comprehensive comparison across all three major MRM categories: generative, semi-scalar, and discriminative.
For \textbf{generative MRMs}, we evaluate both non-critic-trained and critic-trained models. The non-critic-trained generative MRMs include proprietary models, such as GPT-5.2~\cite{gpt52}, Claude-3.7-Sonnet~\cite{claude}, and Gemini-1.5-Pro~\cite{gemini}, as well as open-source models of varying sizes, including Qwen2.5-VL (7B/72B)~\cite{bai2025qwen2}, Llama-3.2 (11B/90B)~\cite{llama}, and InternVL3 (8B/78B)~\cite{zhu2025internvl3}. The critic-trained generative MRMs are further divided into fast-thinking models, such as LLaVA-Critic~\cite{xiong2024llava} and UnifiedReward~\cite{wang2025unified2}, and slow-thinking models, such as R1-Reward~\cite{zhang2025r1}, UnifiedReward-Think~\cite{wang2025unified}, and MR. Judge-7B-SFT-RL~\cite{pi2025mr}. For \textbf{semi-scalar MRMs}, we compare MM-RLHF-Reward~\cite{zhang2025mm}. 
For \textbf{discriminative MRMs}, we consider IXC-2.5-Reward~\cite{zang2025internlm}, Skywork-VL Reward~\cite{wang2025skywork}, BaseReward~\cite{zhang2025basereward}, and Omni-RewardModel-BT~\cite{jin2025omni}.

\subsection{Implementation Details}
\label{subsec:detail}
Our DT2IT-MRM is a discriminative MRM built upon two backbone models, namely Qwen3-VL-8B-Instruct (the default backbone) and Qwen2.5-VL-7B-Instruct, with a linear layer serving as the reward head. 
Our loss function is the Bradley-Terry style loss function (\cref{eq:origin_loss,eq:ourform}), without any auxiliary losses.

For the training data, in Stage 1, we construct 337K single-image preference pairs using our unbiased preference distillation pipeline and 133K multi-image preference pairs using our reformulation method, forming our initial 470K preference dataset $D_0$. 
In Stage 2, we curate five open-source datasets: RLAIF-V~\cite{yu2025rlaif}, VLFeedback~\cite{li2024vlfeedback}, POVID~\cite{zhou2024aligning}, WildVision-Battle~\cite{lu2024wildvision}, and MM-RLHF~\cite{zhang2025mm}. Our final training dataset comprises 929K preference pairs, including curated open-source data and $D_0$.
We also perform decontamination by removing overlapping samples between our training data and benchmarks.

For the training strategy, we perform a grid search over learning rates \{$1e-5$, $3e-6$, $1e-6$, $3e-7$\} and finally select $1e-5$. The model is trained for one epoch with a batch size of 512, a cosine learning rate scheduler, a warmup ratio of 0.1, and a maximum image resolution of $518^2$, using the LLaMA-Factory~\cite{zheng2024llamafactory} framework. 
All training is conducted on 64 Ascend 910B3 64GB NPUs.

\subsection{Results and Analysis}

We evaluate our DT2IT-MRM on the three multimodal reward benchmarks and the results are shown in \Cref{overall_preformance}. Compared with top-performing multimodal reward models (MRMs), DT2IT-MRM achieves the best overall accuracy on both VL-RewardBench and Multimodal RewardBench. Notably, on Multimodal RewardBench, DT2IT-MRM achieves a substantial relative improvement of 8.9\% over the previous SOTA model, BaseReward~\cite{zhang2025basereward}. On MM-RLHF-RewardBench, DT2IT-MRM attains the second-best overall performance.
Considering the results across all three benchmarks, our DT2IT-MRM demonstrates the strongest overall performance, specially in terms of Overall Acc, improving upon the previous best (BaseReward) by 7\%. Notably, BaseReward is trained on 2.8 million samples, whereas DT2IT-MRM uses only 929k, highlighting substantially higher data efficiency.

On VL-RewardBench (\cref{vlrewardbench}), our DT2IT-MRM achieves the best overall performance, with an overall accuracy of 83.5\%, outperforming the previous best model, BaseReward, by 1.1\%. 
Compared with the top-tier proprietary model, GPT-5.2, DT2IT-MRM surpasses it on nearly all metrics, further demonstrating its superiority.
Notably, the reported results of BaseReward suffer from serious issues: its subset scores are clearly inconsistent with the reported overall accuracy. Since BaseReward is not open-sourced and parts of its training data are unavailable, its results cannot be reproduced. 
Similar metric inconsistencies are observed in several other models, whose results we therefore re-evaluated.
In addition, the original reported results of R1-Reward~\cite{zhang2025r1} also exhibit critical flaws. 
During its evaluation, the order of the two responses is not randomized, with the chosen response always preceding the rejected one. Moreover, its result extraction code is unreliable. Extensive repetitive loops are observed in its responses, causing result extraction to fail for many samples.

\cref{multimodal_rewardbench} shows the comparison on the Multimodal RewardBench.
Our DT2IT-MRM ranks first in overall performance, outperforming the previous best open-source model by 4.9\%. We attains the best or second-best results across most evaluation dimensions.
The relatively weaker performance on the knowledge and coding dimensions is mainly due to insufficient world knowledge coverage and the lack of code-related preference pairs in our training data.

On MM-RLHF-RewardBench (\cref{mmrlhf_rewardbench}), DT2IT-MRM attains the second-best results on both overall metrics, Acc and Acc+, trailing the best model by only 0.58\% on Acc. Notably, DT2IT-MRM achieves the best performance on the Short subset, reaching 100\% accuracy. Although DT2IT-MRM does not rank first on this benchmark, it demonstrates more balanced overall performance across multiple benchmarks.

\subsection{Downstream Applications}
\textbf{Inference-time Scaling for MLLMs.} We validate the effectiveness of our DT2IT-MRM under inference time scaling, where MRMs are leveraged during inference to improve MLLM performance. Specifically, we perform best-of-n sampling~\cite{stiennon2020learning} using multiple MRMs based on three different MLLMs. We evaluate on three widely used benchmarks including LLaVA-in-the-Wild~\cite{liu2023visual}, LLaVA-Wilder~\cite{li2024llavanext-strong}, and MM-Vet~\cite{yu2023mm}, covering diverse perception and reasoning tasks. For each benchmark, we first prompt each MLLM to generate four diverse responses per question with a temperature of 0.9, and then use different MRMs to select the best response. As shown in \cref{inference_time_scaling}, our DT2IT-MRM consistently achieves the largest performance gains across different model families, substantially outperforming IXC-2.5-Reward and Skywork-VL Reward under inference-time scaling.

\textbf{Offline Reinforcement Learning for MLLMs.} We further compare the effectiveness of our DT2IT-MRM with that of other baseline MRMs in offline reinforcement learning. Specifically, following LLaVA-Critic~\cite{xiong2024llava}, we conduct one-stage DPO training based on llava-onevision-qwen2-7b-ov-hf, where the multimodal preference datasets are constructed using different MRMs. Again following LLaVA-Critic, we evaluate the models before and after DPO training on three benchmarks: LLaVA-in-the-Wild~\cite{liu2023visual}, LLaVA-Wilder~\cite{li2024llavanext-strong}, and MMHal-Bench~\cite{sun2023aligning}. As shown in \cref{offline_rl}, DPO training with our DT2IT-MRM consistently achieves the best performance across all three benchmarks compared with training using baseline MRMs.

\definecolor{lightblue}{RGB}{230,242,255}
\begin{table*}[!ht]
\centering
\caption{Results on LLaVA-in-the-Wild, LLaVA-Wilder, and MM-Vet for MLLM inference-time scaling. We use the base model to sample four responses for each question and employ our DT2IT-MRM along with two baseline MRMs to select the best response among the four. DT2IT-MRM significantly outperforms the baselines.}
\label{inference_time_scaling}
\setlength{\tabcolsep}{6pt}
\begin{tabular}{l|ccc}
\toprule
\textbf{Model} & \textbf{LLaVA-in-the-Wild} & \textbf{LLaVA-Wilder} & \textbf{MM-Vet} \\
\midrule
InternVL2.5-8B & 72.1 & 63.1 & 51.2  \\
InternVL2.5-8B + IXC-2.5-Reward  & 76.0$^{\uparrow 3.9}$ & 63.7$^{\uparrow 0.6}$ & 52.0$^{\uparrow 0.8}$ \\
InternVL2.5-8B + Skywork-VL Reward  & 76.4$^{\uparrow 4.3}$ & 72.2$^{\uparrow 9.1}$ & 54.6$^{\uparrow 3.4}$ \\
\rowcolor{lightblue}
InternVL2.5-8B + DT2IT-MRM  & 81.8$^{\uparrow 9.7}$ & 74.2$^{\uparrow 11.1}$ & 61.4$^{\uparrow 10.2}$ \\
\midrule
LLaVA-OneVision-7B & 74.6 & 63.4 & 46.2  \\
LLaVA-OneVision-7B + IXC-2.5-Reward  & 75.6$^{\uparrow 1.0}$ & 64.5$^{\uparrow 1.1}$ & 49.6$^{\uparrow 3.4}$ \\
LLaVA-OneVision-7B + Skywork-VL Reward  & 79.4$^{\uparrow 4.8}$ & 73.1$^{\uparrow 9.7}$ & 51.9$^{\uparrow 5.7}$ \\
\rowcolor{lightblue}
LLaVA-OneVision-7B + DT2IT-MRM & 86.6$^{\uparrow 12.0}$ & 75.0$^{\uparrow 11.6}$ &  56.2$^{\uparrow 10.0}$  \\
\midrule
Qwen2.5-VL-7B & 87.3 & 70.6 & 57.8  \\
Qwen2.5-VL-7B + IXC-2.5-Reward  & 88.7$^{\uparrow 1.4}$ & 71.3$^{\uparrow 0.7}$ & 56.8$^{\downarrow 1.0}$ \\
Qwen2.5-VL-7B + Skywork-VL Reward  & 90.5$^{\uparrow 3.2}$ & 78.8$^{\uparrow 8.2}$ & 54.9$^{\downarrow 2.9}$ \\
\rowcolor{lightblue}
Qwen2.5-VL-7B + DT2IT-MRM & 93.3$^{\uparrow 6.0}$ & 79.8$^{\uparrow 9.2}$ & 58.7$^{\uparrow 0.9}$  \\
\bottomrule
\end{tabular}
\end{table*}

\definecolor{lightblue}{RGB}{230,242,255}
\begin{table*}[!ht]
\centering
\caption{Results on LLaVA-in-the-Wild, LLaVA-Wilder, and MMHal-Bench for offline RL using different MRMs.}
\label{offline_rl}
\setlength{\tabcolsep}{6pt}
\begin{tabular}{l|ccc}
\toprule
\textbf{Model} & \textbf{LLaVA-in-the-Wild} & \textbf{LLaVA-Wilder} & \textbf{MMHal-Bench} \\
\midrule
llava-onevision-qwen2-7b-ov-hf & 78.7 &	60.9 & 2.93  \\
llava-onevision-qwen2-7b-ov-hf + IXC-2.5-Reward  & 82.5 & 64.4 & 3.11 \\
llava-onevision-qwen2-7b-ov-hf + Skywork-VL Reward  & 80.6 & 63.1 & 3.04 \\
\rowcolor{lightblue}
llava-onevision-qwen2-7b-ov-hf + DT2IT-MRM  & \textbf{84.6} & \textbf{73.3} & \textbf{3.18} \\
\bottomrule
\end{tabular}
\end{table*}

\subsection{Ablations Study}

\textbf{Debiased Preference Distillation Pipeline.} We first conduct ablation studies to validate the effectiveness of the last three steps in our pipeline. As shown in \cref{ablation_all}, our response diversity enhancement module leads to consistent performance improvements across three benchmarks by introducing more diverse and informative preference signals. Furthermore, pointwise scoring yields more pronounced gains by mitigating position bias and removing unreliable preference pairs constructed by listwise scoring.

\begin{table*}[!ht]
\caption{Ablation study of our two preference data construction methods and iterative training framework.}
\label{ablation_all}
\centering
\setlength{\tabcolsep}{0.75mm}
\renewcommand{\arraystretch}{0.95}
\begin{small}
\begin{tabular}{lcccccc}
\toprule
\multirow{2}{*}{\textbf{Methods}} & \multirow{2}{*}{\textbf{\#Data}} & \multirow{2}{*}{\textbf{VL-RewardBench}} & \textbf{Multimodal} & \textbf{MM-RLHF-} & \textbf{Mean} & \textbf{Overall} \\
 & & & \textbf{RewardBench} & \textbf{RewardBench} & \textbf{Acc} & \textbf{Acc} \\
\midrule
\rowcolor{orange!20}
\multicolumn{7}{l}{\textit{\textbf{Our debiased preference distillation pipeline [Method 1]}}} \\
\midrule
Listwise Scoring & 355K & 74.7 & 76.4 & 77.1 & 76.1 & 76.1 \\
+ Diversity Enhancement  & 405K & 75.6 & 77.2 & 79.4 & 77.4 & 76.9 \\
+ Pointwise Scoring & 337K & \textbf{77.7} & \textbf{78.8} & \textbf{81.2} & \textbf{79.2}  & \textbf{78.7} \\
\midrule
\rowcolor{orange!20}
\multicolumn{7}{l}{\textit{\textbf{Our text-to-image preference reformulation [Method 2]}}} \\
\midrule
Omni-Reward (baseline) & 50K & 46.0 & 55.0 & 47.6 & 49.6 & 53.0  \\
Ours & 50K  & \textbf{56.5}  & \textbf{65.0} & \textbf{59.4} & \textbf{60.3} &  \textbf{63.1} \\
\midrule
\rowcolor{orange!20}
\multicolumn{7}{l}{\textit{\textbf{Our iterative training framework [Method 3]}}} \\
\midrule
Initial data $D_0$& 470K & 79.1 & 79.3 & 84.7 & 81.0 & 79.4 \\
Initial data $D_0$ + uncurated data (baseline) & 942K & 77.9 & 74.2 & 92.9 & 81.7 & 75.5 \\
Our iteration 1 & 869K & 82.1 & \textbf{79.4} & 82.4 & 81.3 & 80.0 \\
Our iteration 2 (Initial data $D_0$ + curated data) & 929K &  \textbf{83.5} & 79.3 & \textbf{89.4} & \textbf{84.1} & \textbf{80.5} \\
\midrule
\rowcolor{orange!20}
\multicolumn{7}{l}{\textit{\textbf{Overall ablation of our three methods}}} \\
\midrule
Method 1 & 337K & 77.7 & 78.8 & 81.2 & 79.2  & 78.7 \\
+Method 2 & 470K & 79.1 & 79.3 & 84.7 & 81.0 & 79.4 \\
+Method 3 & 929K  & \textbf{83.5} & \textbf{79.3} & \textbf{89.4} & \textbf{84.1} & \textbf{80.5} \\
\bottomrule
\end{tabular}
\end{small}
\end{table*}

\begin{table*}[!ht]
\caption{Comparison of our unbiased preference distillation pipeline with existing preference datasets.}
\label{ablation_our_data}
\centering
\setlength{\tabcolsep}{0.75mm}
\renewcommand{\arraystretch}{0.95}
\begin{small}
\begin{tabular}{lccccc}
\toprule
Datasets & VL-RewardBench & Multimodal RewardBench & MM-RLHF-RewardBench & \textbf{Mean Acc} & \textbf{Overall Acc} \\
\midrule
MM-RLHF (50K)  &  51.2 & 75.7 & 94.7 & 73.9 & 71.2  \\
Ours (50K) &  \textbf{75.9} & \textbf{78.0} & 77.1 & \textbf{77.0} & \textbf{77.5}  \\
\midrule
RLAIF-V (83K) & 77.9 & 71.9 & 72.4 & 74.1 & 73.2 \\
Ours (83K)  &  76.3 & \textbf{79.1} & \textbf{76.5} & \textbf{77.3} & \textbf{78.5}  \\
\midrule
VLFeedback (317K)  &  63.4 & 74.9 & 81.8 & 73.4 & 72.7  \\
Ours (317K) & \textbf{77.9}  & \textbf{79.1} & 80.6 & \textbf{79.2} & \textbf{78.9} \\
\midrule
MMPR v1.2 (337K)  &  70.9 & 64.4 & 73.5 & 69.6 & 65.9  \\
Ours (337K) & \textbf{77.7}  & \textbf{78.8} & \textbf{81.2} & \textbf{79.2} & \textbf{78.7} \\
\bottomrule
\end{tabular}
\end{small}
\end{table*}

To further validate the advantages of the data constructed by our pipeline over existing preference datasets, we train MRMs separately using our data and four commonly used open-source preference datasets under the same data volume. Notably, these four datasets cover diverse multimodal preference construction methods. Specifically, to ensure fair comparison, we randomly sample 50K, 83K, and 317K preference pairs from the full dataset constructed by our pipeline (337K in total). For comparison with MMPR v1.2, we randomly sample 337K preference pairs from MMPR v1.2.
As shown in \cref{ablation_our_data}, although our data may slightly underperform some open-source datasets on individual benchmarks, it exhibits clearly superior overall performance across multiple benchmarks, indicating higher data quality.

\textbf{Text-to-image Preference Reformulation.} To validate the superiority of our reformulation method over the baseline method adopted in Omni-Reward~\cite{jin2025omni}, we first randomly sample 50k raw text-to-image preference data from EvalMuse~\cite{han2024evalmuse} and HPDv2~\cite{wu2023human}. We then reconstruct these preferences using the two methods, respectively, and train MRMs on the resulting datasets. 
As shown in \cref{ablation_all}, our method achieves substantial improvements over the baseline, with gains of at least 10\% on each benchmark.

\textbf{Ablation on Iterative Training Framework.} 
To assess the effectiveness of our iterative training framework, we construct a baseline that adopts non-iterative training by directly mixing the initial data $D_0$ from our first stage with five uncurated open-source preference datasets (described in \cref{subsec:detail}). As shown in \cref{ablation_all}, naively incorporating uncurated raw data leads to obvious performance degradation on both VL-RewardBench and Multimodal RewardBench, along with a notable drop in Overall Acc across all three benchmarks. This observation indicates that open-source preference datasets contain considerable noise.
To curate these open-source datasets and further improve MRM performance, our framework iteratively alternates between training MRMs and curating existing preference data. 
In the first iteration, we curate the RLAIF-V, POVID, and VLFeedback datasets, while in the second iteration we curate the WildVision-Battle and MM-RLHF datasets. Across the two iterations, our MRM exhibits consistent performance gains on multiple benchmarks. After curating all five open-source datasets, our method demonstrates substantially better overall performance across three benchmarks, achieving a 5\% gain in overall accuracy compared to the baseline.
\section{Conclusions}

We introduce DT2IT-MRM, which comprises a debiased preference distillation pipeline, a novel text-to-image preference reformulation method, and an iterative training framework. 
Our preference distillation effectively mitigates both textual style bias and positional bias. 
Compared to existing preference data, our constructed data simultaneously exhibit three key characteristics: diverse preference strengths, effective mitigation of textual style bias, and high reliability. 
Additionally, our reformulation method enables more efficient utilization of text-to-image preference data compared to prior approaches. 
Moreover, our iterative training framework progressively improves both MRM performance and data quality by alternately training the reward model and curating preference data. 
Extensive experiments validate the effectiveness of our three core innovations and demonstrate the compelling performance of our DT2IT-MRM.



{
    \small
    \bibliographystyle{ieeenat_fullname}
    \bibliography{main}
}

\newpage
\appendix
\onecolumn




\section{Prompt Collection}
\label{appendix_prompt_collection}
In our debiased preference distillation pipeline, our prompt collection (\cref{tab:prompt_source}) covers three major domains: visual understanding, visual reasoning, and multimodal safety. 
For visual understanding, we consider a diverse set of tasks, including general visual question answering (VQA), OCR, chart understanding, and document understanding.
For visual reasoning, we focus on science and mathematics.
For multimodal safety, the prompts are derived from SPA-VL~\cite{zhang2024spa}, which covers a wide range of harmfulness domains.

\begin{table*}[!ht]
\centering
\caption{Overview of the prompt source of our debiased preference distillation pipeline.
}
\label{tab:prompt_source}
\begin{tabularx}{\textwidth}{>{\raggedright\arraybackslash}p{3cm}|X}
\hline
\textbf{Category} & \textbf{Dataset} \\
\hline
Visual Understanding & GQA~\cite{hudson2019gqa}, IconQA~\cite{lu2021iconqa}, InfoVQA~\cite{mathew2022infographicvqa}, ChartQA~\cite{masry2022chartqa}, DVQA~\cite{kafle2018dvqa}, MapQA~\cite{chang2022mapqa} \\
\hline
Visual Reasoning & AI2D~\cite{kembhavi2016diagram}, M3CoT~\cite{chen2024m}, ScienceQA~\cite{lu2022learn}, MMK12~\cite{meng2025mm}, CLEVR-Math~\cite{johnson2017clevr}, GeoQA+~\cite{cao2022augmented}, Geometry3K~\cite{lu2021inter}, Geo170K~\cite{gao2023g}, GEOS~\cite{seo2015solving}, UniGeo~\cite{chen2022unigeo} \\
\hline
Multimodal Safety & SPA-VL~\cite{zhang2024spa} \\
\hline
\end{tabularx}
\end{table*}

\section{Evaluation Prompt}
\begin{table*}[!htbp]
\centering
\caption{Evaluation prompt template used for listwise scoring.}
\begin{tcolorbox}[
    arc=4pt, 
    boxrule=1pt, 
    colback=gray!10, 
    colframe=black, 
    boxsep=0pt,
    left=4pt,
    right=4pt,
    width=\linewidth,
]
\label{tab:evalution_prompt}
Please act as an impartial evaluator and assess the quality of several AI assistants' responses to the same user's prompt based on the provided image. Your task is to analyze each response and provide feedback by comparing it with the standard human-generated answer.\\
\#\#\# Evaluation Criteria (score each from 0-10, integer only):\\
1. Accuracy: Accuracy is determined by the number of hallucinations in the response. Hallucinations refer to information that conflicts with or is not present in the image or user's prompt. To detect hallucinations, you must identify any non-existent or inaccurate information by evaluating the following: Does the response include objects or details not present or implied in the image? Are all parts of the response consistent with the image content? Are the visual elements (for example: environment, objects, relationships among objects, and object attributes, including actions, positions, colors, sizes, quantities, identities, ages, etc.) in the image accurately described? Do the metaphorical descriptions (for example: atmosphere, viewing experience, conveyed meaning, purpose, etc.) align with the image?\\
2. Helpfulness: Helpfulness means the answer correctly responds to the user's prompt or follows user's instructions. Are all parts of the response closely related to and appropriate for the user's prompt and the given image? If the user's prompt is ambiguous or has more than one interpretation, does the assistant appropriately request clarification or make reasonable assumptions?\\
3. Completeness: Does the response comprehensively address all the requests in the user's prompt? Does the response include all necessary detailed information? Is the necessary analysis process and analysis perspective complete in the response?\\
4. Language Quality: Is the response fluent, clear, and to the point, avoiding excessive verbosity or redundant content? Does the tone and style align with the task requirements (e.g., academic, casual, explanatory)?\\
5. Creativity: Consider the creativity and novelty of the response when needed, such as writing a story. Is the additional detailed information beyond the user's question in the response reasonable and rich? Does the response demonstrate creativity or unique insights? If the user's question does not require creativity, then score 10.\\
6. Ethics: Please evaluate four aspects: safety, privacy, fairness and harmfulness. For safety, assess if the response gives appropriate warnings or avoids providing advice on sensitive topics, such as medical images. For privacy, does the assistant refrain from stating identification information in the image that could compromise personal privacy? For fairness, assess if the response treats individuals and communities fairly, avoiding biases. For harmfulness, identify if the response contains content that may potentially incite violence, be classified as NSFW (Not Safe For Work), or involve other unmentioned ethical considerations. Consider any content that could be deemed offensive, inappropriate, or ethically problematic beyond the explicitly listed criteria. If the user's question doesn't need ethical considerations, then score 10.\\
\#\#\# Scoring:\\
- Each criterion: integer 0-10 (0 = extremely poor, 10 = excellent).\\
- Overall Score: The overall score should be calculated by taking the weighted average of the scores across all evaluation criteria, expressed as a decimal with up to 0.01 precision.\\
\#\#\# Important Instructions:\\
- During your evaluation, first generate your own complete reference answer to the user's question. Note that the standard answer may contain errors; please examine the image carefully and respond. Next, assign weights to all criteria based on their relative importance. All weights must sum to 1.0. Then, compare each assistant's response with the factual information in the image, the user's prompt, and your reference answer. Compare the assistants' responses and assign a higher Overall Score to the better one.\\
- The order in which all responses are presented is not related to the quality of the responses.\\
- The length of a response does not directly correlate with its quality. A longer response is not necessarily better.\\
\#\#\# User Prompt:\\
\{question\}\\
\#\#\# Standard Human-Generated Answer:\\
\{gt\}\\

\{evaluated\_answers\}
\end{tcolorbox}
\label{prompt}
\end{table*}

\end{document}